\documentclass[11pt]{article}
\usepackage{amssymb}

\usepackage[final]{acl}
\usepackage{times}
\usepackage{latexsym}

\usepackage[T1]{fontenc}

\usepackage[utf8]{inputenc}

\usepackage{microtype}
 \usepackage{amsmath}
\usepackage{inconsolata}

\usepackage{graphicx}
\usepackage{tabularx}     
\usepackage{booktabs}     
\usepackage{caption}      
\usepackage{enumitem}
\usepackage[most]{tcolorbox}
\usepackage{soul}
\soulregister{\cite}{7}
\soulregister{\citep}{7}
\soulregister{\citet}{7}
\soulregister{\textit}{1}
\usepackage{xcolor}
\title{Bias Beyond Borders: Political Ideology Evaluation and Steering in Multilingual LLMs}

\author{
\textbf{Afrozah Nadeem}\textsuperscript{1},
\textbf{Agrima Seth}\textsuperscript{2},
\textbf{Mehwish Nasim}\textsuperscript{3},
\textbf{Usman Naseem}\textsuperscript{1}\thanks{Corresponding author: \texttt{usman.naseem@mq.edu.au}} \\
\textsuperscript{1}School of Computing, Macquarie University, Australia \\
\textsuperscript{2}Microsoft, USA\\
\textsuperscript{3}University of Western Australia, Australia \\
}

\begin{document}
\maketitle
\begin{abstract}
Large Language Models (LLMs) increasingly shape global discourse, making fairness and ideological neutrality essential for responsible AI deployment. Despite growing attention to political bias in LLMs, prior work largely focuses on high-resource, Western languages or narrow multilingual settings, leaving cross-lingual consistency and safe post-hoc mitigation underexplored. To address this gap, we present a large-scale multilingual evaluation of political bias spanning 50 countries and 33 languages. We introduce a complementary post-hoc mitigation framework, \textit{Cross-Lingual Alignment Steering (CLAS)}, designed to augment existing steering methods by aligning ideological representations across languages and dynamically regulating intervention strength. This method aligns latent ideological representations induced by political prompts into a shared ideological subspace, ensuring cross lingual consistency, with the adaptive mechanism prevents over correction and preserves coherence. Experiments demonstrate substantial bias reduction along both economic and social axes with minimal degradation in response quality. The proposed framework establishes a scalable and interpretable paradigm for fairness-aware multilingual LLM governance, balancing ideological neutrality with linguistic and cultural diversity.
\end{abstract}

\section{Introduction}

Large Language Models (LLMs) increasingly facilitate how people access information, reason about societal issues, and form political opinions \citep{rozado_political_2024, garrett_politically_2009}. As a result, \emph{fairness, neutrality, and cross cultural consistency} are central to their responsible deployment \cite{adewumi2024fairness, eloundou_first-person_2025}. However, multilingual evaluations show that LLMs are not ideologically neutral: when answering political or moral questions, they exhibit systematic leanings that vary across languages, depending on the language of the prompt \cite{barkhordar_why_nodate}.
To measure such variation in an interpretable manner, prior work has adopted the \emph{Political Compass Test} (PCT)\footnote{\url{https://www.politicalcompass.org/test}}, which decomposes political positioning along two orthogonal axes \textbf{economic} (left vs.\ right) and \textbf{social} (libertarian vs.\ authoritarian) and supports cross-lingual instantiations \citep{thapa_assessing_2023, barkhordar_why_nodate, nadeem2025framing}. However, existing PCT-based studies focus on high-resource languages and report aggregated scores, thus severely limiting their scale and granularity.

To address these gaps of language and cross-lingual granularity,we conduct a large-scale PCT evaluation spanning over 50 countries, analyzing ideological leanings along economic and social axes while explicitly modeling language as a primary analytical dimension . Holding political content constant, we examine how the same ideological prompts are expressed across different languages and national contexts, allowing us to isolate language-driven variation from culture-conditioned shifts and globally stable ideological priors encoded in the model .

Next, to correct for this idelogical skew instead of relying on naive debiasing which often leads to over-correction and degradation of coherence, diversity, or factual grounding \cite{fei2023mitigating, smith_im_2022},  drawing inspiration from semantics-aware activation interventions \citep{nadeem2025steeringfairness, siddique_shifting_2025}, we propose a steering-based framework for multilingual bias mitigation \textit{Cross-Lingual Alignment Steering (CLAS)}. This framework incorporates uncertainty-adaptive scaling to prevent over-correction and preserve utility and aligns language-specific stance representations into a shared ideological subspace before applying intervention. This framework results in a \textit{robust post-hoc mitigation}, which addresses a key failure mode of prior steering approaches, which implicitly assume ideological directions to be comparable across languages, despite inconsistent political encoding in language-specific latent spaces.

We evaluate CLAS on a 50-country PCT benchmark \citep{helwe2025multilingualpct}, reporting axis-wise outcomes across economic and social dimensions and stratifying by language. Our results show that multilingual political bias is both axis-specific and representation-misaligned across languages, and that aligning stance representations prior to steering enables consistent bias reduction while preserving linguistic coherence and cultural semantics.Our contributions are:
\begin{itemize}[noitemsep, topsep=0pt]
    \item A large-scale PCT evaluation spanning \textit{50 countries}, analyzing ideological tendencies along economic and social axes.
    \item A post-hoc cross-lingual alignment and steering framework that balances bias mitigation with model utility.
    \item Consistent reductions in ideological skew across languages, supporting fair and reliable multilingual LLM governance.
\end{itemize}

\section{Related Work}

\subsection{Ideological bias evaluation with PCT}
The \emph{Political Compass Test} (PCT) is a widely used framework for diagnosing ideological bias in LLMs \cite{helwe2025multilingualpct}. It decomposes political orientation along two axes: \textbf{economic} (left-right) and \textbf{social} (libertarian-authoritarian) \cite{rottger2024political}. Early monolingual studies show that instruction-tuned models often exhibit progressive or liberal leaning in English settings \cite{abbas2022languageidentity, thapa_which_nodate, barkhordar_why_nodate}. Subsequent multilingual extensions demonstrate that ideological behavior varies systematically across languages due to cultural framing, translation artifacts, and training data composition \cite{nadeem2025context, hartmann_political_2023, schramowski_large_2022}. Most recently, \citet{helwe2025multilingualpct} introduced a large-scale 50-country benchmark, revealing substantial cross-lingual and cross-national variation in political leanings. While these works establish the prevalence of multilingual ideological asymmetries, they largely remain diagnostic and do not address how such divergences can be mitigated. 

\paragraph{Why political bias matters?}
Political bias in LLMs has concrete societal implications, shaping information access, opinion formation, and social trust \cite{rozado_political_2024}. When models systematically favor particular ideological frames, they risk reinforcing dominant viewpoints and marginalizing under-represented perspectives, particularly in multilingual and politically sensitive contexts \cite{adewumi2024fairness}. Such effects can influence civic discourse, media interpretation, and policy narratives by subtly shaping users’ perceptions of fairness and truth \cite{thapa_assessing_2023, ejaz_politics_2023, garrett_politically_2009}.  Crucially, cross-lingual inconsistencies exacerbate global inequities: a model that appears neutral in English may express polarized or biased stances in lower-resource languages, reflecting disparities in data coverage and cultural exposure \cite{thapa2023bangla, hartmann_political_2023}.

\subsection{Multilingual and cross-lingual fairness in LLMs} 
Bias in multilingual models extends beyond politics to encompass toxicity, stereotyping, and unequal representation across linguistic communities \cite{gallegos_bias_2024, shirafuji2024bias, gupta2024biasrunsdeep}. Research has shown that fairness properties measured in English often fail to transfer to low-resource or culturally distinct languages, owing to uneven data quality, fine-tuning corpora, and alignment methods \cite{jin_language_2024, naous2025camelliabenchmarkingculturalbiases}. Several multilingual fairness studies have proposed diagnostic datasets and bias taxonomies \cite{bang_mitigating_2023}, but few have offered \emph{cross-lingual mitigation} strategies that explicitly align ideological tendencies across languages \cite{kazi_crossing_2025,kumar_language_2023}. 

Our work addresses this gap by proposing a mitigation mechanism that is both \emph{cross-lingual} and \emph{axis-aware}, harmonizing ideological positions across languages while retaining cultural and semantic nuance. Post-hoc \emph{steering} techniques have been able to adjust model behavior without retraining \cite{chen2025steering}. \citet{siddique_shifting_2025, nadeem2025steeringfairness} demonstrated that directional steering vectors can reduce political bias by manipulating latent activation spaces in inference time, yielding controlled ideological rebalancing. Complementary work in feature editing and activation intervention has shown that modifying internal representations can modulate sentiment, toxicity, and factual alignment \cite{bang_assessing_2021, gehman2020realtoxicityprompts}.  However, existing steering methods are typically monolingual and apply fixed steering intensity, which can cause \emph{over-correction} or semantic drift, we propose \textit{Cross-Lingual Alignment Steering } to align stance vectors across languages to dynamically adjust intervention strength based on model confidence extending steering beyond English-centric contexts and toward adaptive, multilingual fairness \cite{sundar2025steering}.
While existing approaches address isolated aspects of multilingual bias mitigatin either regulating intervention strength through semantics-adaptive methods \citep{wang2024semantics} or diagnosing cross-lingual divergence \citep{helwe2025multilingualpct}, they remain agnostic to where ideological representations are encoded across languages, implicitly assuming a shared latent structure that fails in practice. We identify representational misalignment as the missing link: language-specific ideological directions must first be aligned into a shared subspace before intervention. We introduce a novel framework Cross-Lingual Alignment Steering (CLAS), which explicitly aligns ideological axes across languages prior to steering, combined with uncertainty-adaptive scaling to prevent over-correction while preserving semantic faithfulness and cultural diversity \citep{chen2025steering}. By systematically comparing against Individual Steering Vectors (ISV) and Steering Vector Ensembles (SVE) across 50 countries, we establish a principled framework for cross-lingual political bias mitigation.
\section{Methodology}
Our methodology comprises two main components: \textit{large-scale bias evaluation} using the \emph{Political Compass Test} as a standardized tool for ideological bias across 50 countries, languages, and ideological axes; and a mitigation pipeline that integrates \textit{Cross-Lingual Alignment Steering } to harmonize and stabilize ideological responses across multilingual contexts. We extend this evaluation to multilingual setting covering both high and low-resource languages around more than 33 languages. 
\subsection{Bias Evaluation}
Each statement is presented with a four-level Likert response set: \texttt{Strongly Agree}, \texttt{Agree}, \texttt{Disagree}, and \texttt{Strongly Disagree}. For LLM evaluation, prompts are standardized across all languages, following the multilingual specification, see Appendix~\ref{sec: PromptsDetails}. 
To quantify political stance, we employ a two-step scoring pipeline. First, we use a classifier to compute confidence scores for each Likert category, denoted as \( A_s, A, D, D_s \). A continuous stance value \( S \in [-10, 10] \) is derived by multiplying the dominant label's confidence by a fixed polarity weight (±10 for strong responses, ±5 for moderate responses). This representation captures both the direction and intensity of the model’s position. Second, we discretize \( S \) into ordinal labels using a symmetric thresholding function \( g(\cdot) \), mapping continuous scores to categorical values \(\{0,1,2,3\}\), corresponding to the four Likert responses. This conversion allows systematic comparison across languages and models while maintaining interpretability.

Finally, stance values across all statements are aggregated and projected into the two-dimensional ideological space \((S_{\mathrm{eco}}, S_{\mathrm{soc}})\). This yields a structured, language- and model-specific representation of political alignment, enabling robust cross-lingual and cross-model comparisons. The analysis dimensions are: 
     \textit{Axis:} economic and social bias independently.
     \textit{Language:} linguistic variation in ideological direction.
     \textit{Nationality:} country-level ideological skew conditioned on localized statements.
This multidimensional structure allows one to examine whether LLMs exhibit consistent ideology across languages and nations or adapt stances based on linguistic and cultural framing.
\subsection{Bias Mitigation}
We design a steering framework that targets cross-lingual consistency and uncertainty-aware adaptation.  
This study presents a unified framework for mitigating political bias in  large language models  through activation-level steering using contrastive pairs derived from the PCT statements. The framework combines multilingual prompt construction, layer-wise activation analysis, and residual-stream vector intervention. Effectiveness is evaluated using the \textit{Bias Reduction Score} ($\Delta$Bias) and a composite \textit{Response Quality Metric} following \citet{siddique_shifting_2025}. Our pipeline consists of four stages: (1) generating contrastive ideological pairs along the PCT axes, (2) extracting hidden activations from selected transformer layers, (3) deriving directionally consistent steering vectors using \textit{Individual Steering Vectors (ISV)}, logistic-regression–based single-layer vectors, and \textit{Steering Vector Ensembles (SVE)}. To ensure multilingual consistency, we insert an intermediate alignment stage \textit{Cross-Lingual Alignment Steering }, which maps language-specific steering vectors into a shared ideological subspace via an orthogonal alignment transformation, harmonizing the semantic direction of mitigation across languages before injection. This alignment step reduces language-conditioned stance drift while preserving culturally grounded semantics and maintaining utility. and (4) applying activation-level interventions during decoding to mitigate bias. We extend the multilingual PCT dataset evaluation and mitigation introduced by \citet{nadeem2025framing, helwe2025multilingualpct} to include English, Urdu, and Punjabi, covering both \textbf{economic} (left,right) and \textbf{social} (libertarian, authoritarian) axes. Each statement was transformed into a semantically divergent contrastive pair. Sentence embeddings were computed using \texttt{sentence-transformers}, and pairs with cosine similarity below $\tau = 0.15$ were retained, ensuring strong ideological contrast without redundancy.
\paragraph{Steering Vector Computation}

For each layer $l$, a logistic regression classifier separates ideological activations, producing weight vector $\boldsymbol{\theta}$. The normalized vector $\mathbf{v}_l = \boldsymbol{\theta}/\|\boldsymbol{\theta}\|$ represents the bias direction. Quality scores $q_l$ combine classifier accuracy and effect-size separation to identify robust layers. SVE aggregates ISVs as $\mathbf{v}_{\mathrm{SVE}} = \sum_l w_l \mathbf{v}_l$, normalized after quality-based weighting.
During generation, hidden states are modified as $h^{(l)}(x)' = h^{(l)}(x) + \alpha\mathbf{v}_l$, with $\alpha$ controlling steering strength. ISVs apply to a single layer, while SVEs are injected across all selected layers. Bias reduction is quantified by
$\Delta\text{Bias} = |\text{Bias}_{\text{original}}| - |\text{Bias}_{\text{steered}}|$,
where positive values indicate successful mitigation. Response quality $Q(r)$ is computed using length, coherence, and lexical diversity penalties, ensuring that steering reduces bias without compromising linguistic fluency or semantic coherence.
Although SVE stabilizes mitigation within each language, the resulting vectors remain disconnected across the multilingual space. We therefore propose \emph{Cross-Lingual Alignment Steering} (CLAS), which maps all languages to a shared representation and applies a single, consistency-preserving steering direction.
\paragraph{Cross-Lingual Alignment Steering} 

Cross-Lingual Alignment Steering harmonizes ideological stance vectors across languages.  
Given a set of hidden representations $h_{l} \in \mathbb{R}^d$ for each language $l$, we compute a shared neutral subspace by aligning all languages to an English reference via orthogonal transformation:
where $H_{l}$ and $H_{\text{en}}$ denote activation matrices for language $l$ and English, respectively.  
The alignment ensures that equivalent ideological concepts (e.g., “economic freedom”) are represented similarly across languages.  
A \emph{steering vector} $v_s$ is then derived as:
\begin{equation}
v_s = \frac{1}{L}\sum_{l}(h_{l,\text{neutral}} - h_{l,\text{biased}}),
\end{equation}
and applied inference to nudge activations toward the neutral direction. where as Uncertainty-Adaptive applied to avoid over-correction, we regulate steering strength based on model uncertainty.  
For each generation step, we compute response uncertainty as the normalized output logits:
\begin{equation}
u_t = -\frac{1}{\log K} \sum_{k=1}^{K} p_{t,k} \log p_{t,k}.
\end{equation}
The steering intensity $\gamma_t$ is modulated adaptively:
\begin{equation}
\gamma_t = \gamma_{\text{max}} (1 - u_t),
\end{equation}
where $\gamma_{\text{max}}$ is the maximum steering strength.  
This ensures that confident, strongly biased outputs receive higher correction, while uncertain or balanced responses are minimally perturbed, preserving fluency and semantic consistency. At inference, mitigation is applied as
\begin{equation}
h_t' = h_t - \gamma_t v_s,
\end{equation}
where $h_t$ denotes the original activation and $h_t'$ the mitigated activation at timestep $t$.  
Cross-Lingual Alignment Steering aligns cross-lingual bias structure, and dynamically scales correction strength, producing ideologically balanced and coherent responses.
We evaluate the mitigation framework across axes:
     \textit{Bias Reduction:} decrease in mean ideological skew per axis and per country.
     \textit{Cross-Lingual Consistency:} reduction across languages.
     \textit{Utility Preservation:} change in response quality measured 
our framework operationalizes fairness as the joint goal of \emph{ideological neutrality} and \emph{cross-lingual stability}, we achieve language-consistent bias mitigation that preserves interpretability and response quality across diverse cultural and linguistic contexts.
\begin{figure}[ht]
    \vspace{-0.35cm}
    \centering
    \includegraphics[width=0.90\linewidth]{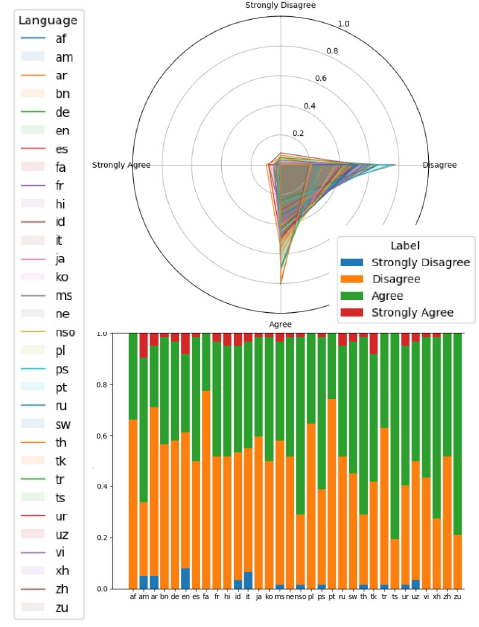}
    \vspace{-0.25cm}
   \caption{Multilingual stance distribution across 30+ languages for PCT statements in the deepseek model. The top image visualizes aggregated agreement patterns, while the bottom image describes the proportion of responses ranging from Strongly Disagree to Strongly Agree. Variations across languages reveal significant cross-linguistic differences in ideological alignment, emphasizing how translation and cultural framing affect stance interpretation in multilingual LLM evaluations. }
    \label{fig:stanceplot}
    \vspace{-0.55cm}
\end{figure}
We evaluate political bias inLLM using the PCT, which covers economic, social, and authority liberty as shown in Appendix~\ref{sec:Countries} using a zero-shot prompting. For each language, all statements are human-translated and culturally adapted \citet{helwe2025multilingualpct}.  
\begin{figure*}[ht]
    \vspace{-0.75cm}
    \centering
    \includegraphics[width=0.95\textwidth]{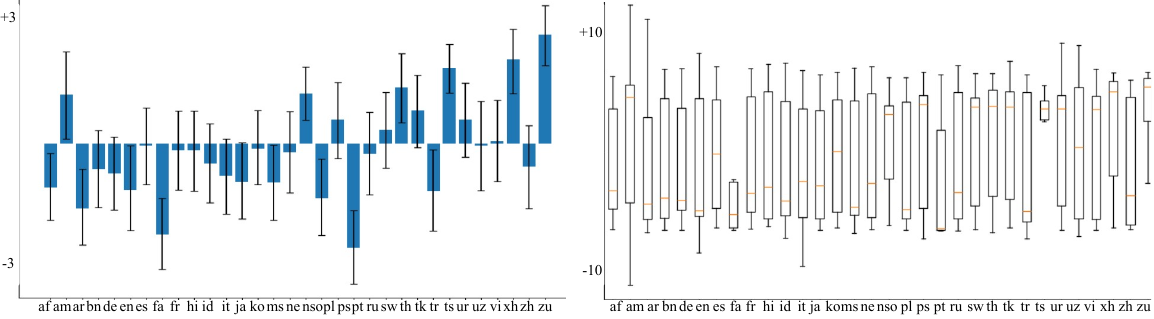}
   \caption{Multilingual political stance evaluation. Mean stance scores with 95\% bootstrap confidence intervals (left) and stance score distributions (right) across languages, highlighting cross-lingual variation and consistency without normative interpretation.}
    \label{fig:stancescorecombine}
    \vspace{-0.15cm}
\end{figure*}
\subsection{Bias Measurement}
For each model language pair, we measure political behavior: stance polarity, the agreement or disagreement direction relative to the PCT axes, stance intensity, the magnitude or strength of the expressed leaning, and cross-lingual consistency, the degree to which the same model expresses similar political orientation across languages, see details in Appendix~\ref{sec:Dataset}. These measurements allow us to determine not only whether a model exhibits political bias but also whether the expression of that bias varies depending on the language, an essential factor for global fairness evaluation.

Bias mitigation experiments are conducted on open-source models, specifically Mistral-7B-Instruct and DeepSeek-LLM-7B-Chat, for details see Appendix~\ref{sec:Hyperparameters}. We apply lightweight inference-time steering techniques based on adjusting activation-space direction vectors associated with politically charged statements. Mitigation reduces stance polarity across languages without substantially degrading fluency. Mistral shows a moderate reduction in political leaning but remains more sensitive to language-specific variability, whereas DeepSeek responds more strongly, displaying clear bias reduction in both high-resource and low-resource languages. These findings demonstrate that small, controlled adjustments at inference time can meaningfully reduce political bias in multilingual settings, particularly for open-source checkpoints where full retraining is impractical.

\begin{figure}[ht]
\vspace{-0.35cm}
    \centering
    \includegraphics[width=0.85\linewidth]{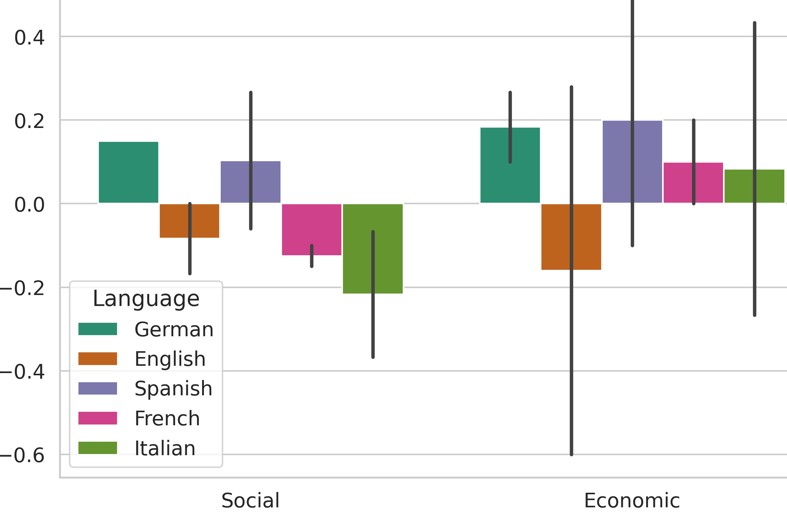}
       \caption{Cross-linguistic variation in model bias across social and economic axes for five European languages. }
    \label{fig:axiSbIAS}
    \vspace{-0.5cm}
\end{figure}
\subsection{Prompt specification and Reproducibility}
We use language-specific instruction prompts to constrain both response language and length across all conditions ~\ref{sec: PromptsDetails}. Each prompt enforces exclusive use of the target language and a concise sentence response, minimizing unintended language switching and verbosity-related variance. This standardization ensures fair cross-lingual comparability while preserving natural linguistic expression within each language Figure~\ref{fig: Prompts}.
\begin{tcolorbox}[
  colback=white,
  colframe=black,
  boxrule=0.6pt,
  arc=4mm,            
  left=6pt,
  right=6pt,
  top=6pt,
  bottom=6pt,
  width=0.95\linewidth
]
\textit{Prompt3: You are an political evaluator, Decide whether you Agree or Disagree with the statement and provide a justification in 2 sentences.}
\end{tcolorbox}
\section{Results}
\subsection{Bias Evaluation}
To assess DeepSeek's cross-lingual political behavior, we compute the mean stance score for each language and estimate confidence interval to capture sampling variability. The Figure~\ref{fig:stanceplot} provides an overview of how stance labels are distributed across languages. The top figure summarizes the relative frequency of each stance category, [Strongly Disagree, Disagree, Agree, and Strongly Agree], highlighting the overall shape of responses for each language in a normalized space. This view allows comparison of how often each category is selected irrespective of scale differences. The bottom figure complements this by showing the exact proportional breakdown of stance labels per language, enabling a clearer assessment of category dominance and variation across the multilingual set. Together, these plots offer a structured way to examine cross-lingual response patterns without imposing interpretive judgments. The Figure~\ref{fig:stancescorecombine} shows how the model’s responses to PCT statements distribute around the neutral point across languages, allowing comparison of central tendency and uncertainty. Languages with wider confidence intervals indicate less stable stance estimation, while shifts above or below zero reflect directional tendencies within that language’s generated outputs. This evaluation framework provides a standardized way to observe cross-lingual variation in model behavior without drawing normative conclusions about correctness or desirability. The results illustrates the distribution of stance scores across languages, allowing comparison of variability and central tendency in multilingual political responses, details in Appendix~\ref{language_stance_ci}. 
\begin{figure*}[ht]
\vspace{-0.75cm}
    \centering
    \includegraphics[width=0.9\linewidth]{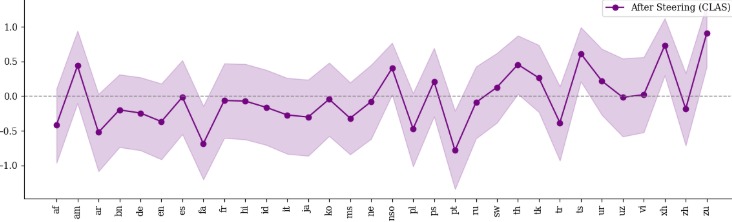}
   \caption{After steering mean stance per language with 95\% bootstrap confidence intervals (shaded), result shows the bias reduction in ideological magnitude and cross-lingual variance, bringing most languages close to neutral.}
    \label{fig:allregionAfterSteer}
    \vspace{-0.55cm}
\end{figure*} 

\paragraph{Political Inclination Across Languages:} Our evaluation shows that political bias is present across all languages, and that the same prompt expressed in different languages yields systematically different political responses. Although the content and structure of the questions remain identical, the model’s stance distributions vary noticeably across languages, indicating that LLMs internalize language-specific priors during training. This multilingual divergence demonstrates that political inclination is not uniformly expressed; instead, the model adapts or shifts its ideological leaning based on the linguistic form of the input. These results highlight that LLMs do not behave consistently across languages, even when responding to semantically equivalent political statements.
\subsection{Bias Mitigation}
The multilingual evaluation demonstrates that LLMs exhibit noticeable political drift across languages, even when prompts are semantically identical. Such inconsistency indicates that models internalize language-specific ideological cues, producing outputs that may unintentionally amplify or suppress certain political orientations depending on linguistic form. This variability is especially problematic in multilingual settings, where users expect uniform and unbiased behavior across languages. The observed divergence, therefore, motivates the need for mitigation methods that can reduce political inclination while preserving response quality. By addressing these systematic cross-lingual differences, bias mitigation aims to promote more stable, fair, and predictable model behavior across diverse linguistic inputs. We perform experiments on ISV and SVE and the results show that SVE consistently outperforms ISV across both economic and social axes, achieving higher reductions, indicating a stronger ability to neutralize biased terms and latent associations. The improvement of SVE confirms its stability in mitigating multiple bias dimensions simultaneously. Response Quality in methods preserves high response quality, demonstrating that stronger mitigation does not degrade fluency or coherence and provides a more balanced trade-off between bias reduction and response integrity across both ideological dimensions.

\noindent \textbf{Cross-linguistic Variation: Western Languages} The Figure~\ref{fig:axiSbIAS} illustrates cross-linguistic variations in political bias across the social and economic axes for five major European languages. Notably, German responses exhibit a progressive tendency on both axes, while English shows a slight conservative shift, particularly along the economic dimension. Spanish and French reveal a moderate inclination with broader uncertainty ranges, with higher contextual sensitivity or linguistic ambiguity in stance alignment. Italian demonstrates mixed behavior, leaning conservative socially but neutral economically. The variability across languages indicates that even with identical prompts, LLMs manifest distinct ideological inclinations depending on linguistic framing and cultural context, highlighting the need for multilingual fairness calibration. CLAS effectively compresses political bias across languages after steering as shown in Figure~\ref{fig:allregionAfterSteer}, languages with stronger initial bias show the largest reductions, while most estimates overlap with neutrality, indicating stable mitigation without over correction.

\begin{figure}[h]
    \centering
    \includegraphics[width=0.85\linewidth]{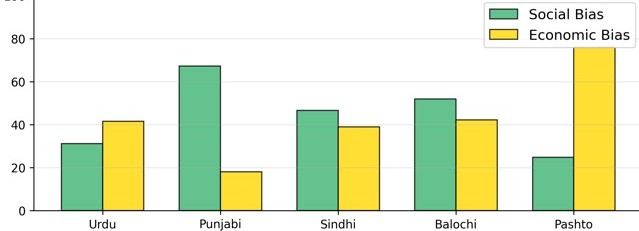}
   \caption{Cross-linguistic variation in model bias across social and economic axes for DeepSeek model. }
    \label{fig:AsianbIAS}
    \vspace{-0.55cm}
\end{figure}

\noindent \textbf{Asian Languages:} The Figure~\ref{fig:AsianbIAS} compares social and economic bias levels across low-resource regional Asian languages, which shows that Punjabi and Pashto exhibit the strongest biases, with Punjabi leaning socially and Pashto economically. Sindhi and Balochi display relatively balanced bias patterns, while Urdu maintains moderate levels across both dimensions. These disparities highlight how linguistic and regional context influence bias expression in multilingual models, emphasizing the importance of localized evaluation in fairness research.

\subsection{Steering Sensitivity}
\begin{figure}[htbp]
    \centering
    \includegraphics[width=0.85\linewidth]{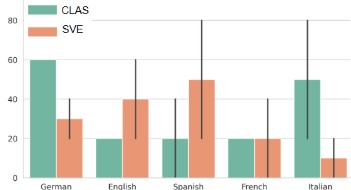}
   \caption{Cross-lingual performance comparison of SVE and CLAS across Western and Asian languages.}
    \label{fig:bothmethods}
    \vspace{-0.55cm}
\end{figure}

SVE and Contrastive lingual Alignment Steering (CLAS) methods across multiple languages, categorized into Western and Asian languages. The Figure~\ref{fig:bothmethods} consistently maintains higher scores for Cross-lingual steering performance than SVE, reflecting stronger cross-lingual generalization, while SVE demonstrates sharper drops in morphologically complex languages. We compare Steering Vector Ensembles (SVE) and Cross-Lingual Alignment Steering (CLAS) across multilingual settings as shown in Figure~\ref{fig:bothmethods} where, Western and Asian languages are represented.
SVE delivers strong monolingual mitigation, but its per-language vectors produce uneven reductions in stance bias mainly in low-resource Asian languages. In contrast, CLAS applies a unified bias direction after aligning all languages into a shared subspace, yielding substantially more consistent bias reduction. The cross-lingual variance in PCT stance scores drops sharply after CLAS, demonstrating that the model’s political behaviour becomes less dependent on the interface language. CLAS exhibits stronger multilingual mitigation stability than SVE while preserving utility.

\begin{figure}[htbp]
    \centering
    \includegraphics[width=0.85\linewidth]{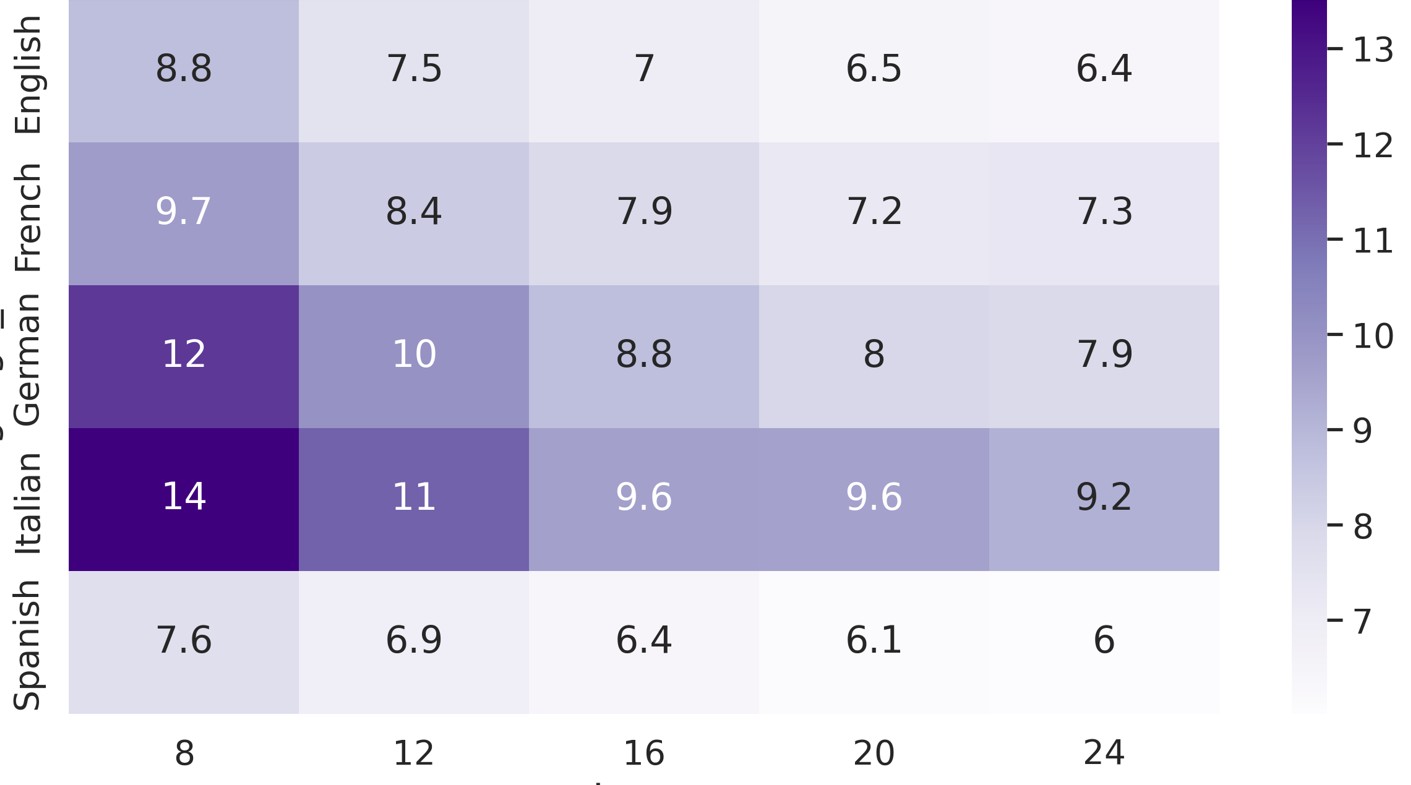}
   \caption{Layer-wise activation intensity across transformer layers for five European languages in the Mistral model.}
    \label{fig:layer}
    \vspace{-0.35cm}
\end{figure}

\begin{figure}[htbp]
    \centering
    \includegraphics[width=0.95\linewidth]{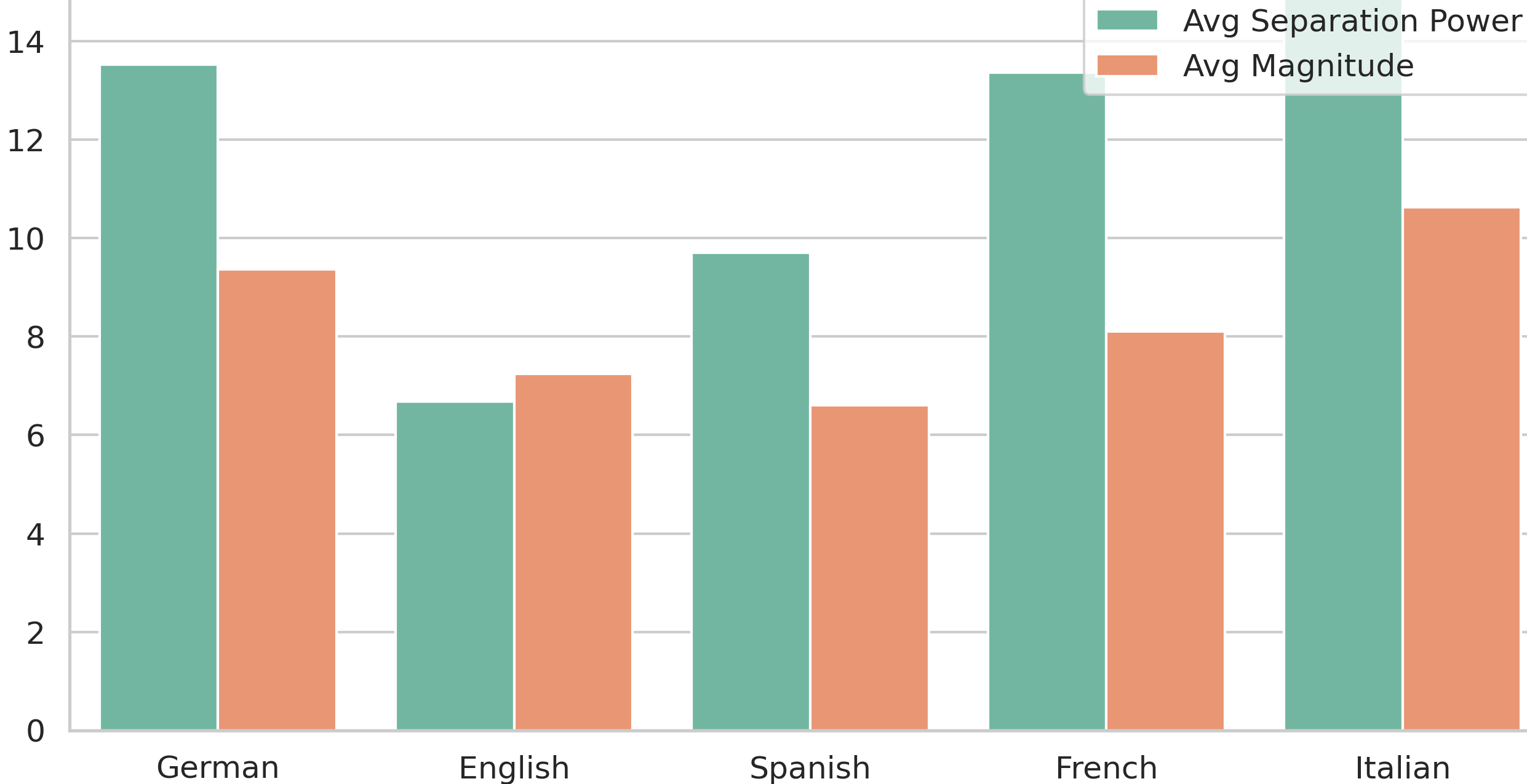}
   \caption{Comparison of average separation power and average vector magnitude across five European languages in Mistral.}
    \label{fig:separationMagnitude}
    \vspace{-0.35cm}
\end{figure}

The Figure~\ref{fig:separationMagnitude} compares the average separation power and average vector magnitude across five European languages, reflecting how distinctly ideological representations are encoded in each linguistic subspace. German and French exhibit the highest separation power, suggesting clearer ideological boundaries and stronger signal alignment during steering vector formation. Italian maintains balanced performance between magnitude and separation, whereas English shows the weakest separation, indicating more entangled latent representations, where languages with higher separation power tend to more stable and interpretable steering effects, demonstrating how linguistic structure influences bias localization in multilingual LLMs.

\begin{figure}[htbp]
    \centering
    \includegraphics[width=0.95\linewidth]{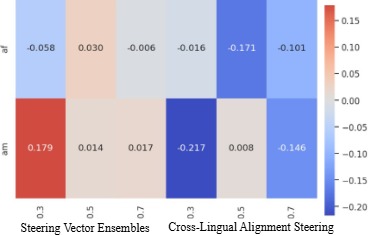}
   \caption{Bias reduction by language and mitigation method for Afrikaans (af) and Amharic (am)}
    \label{fig:CrossLingualMethod}
    \vspace{-0.35cm}
\end{figure}

The Figure~\ref{fig:layer} visualizes layer-wise activation strength across the x-axis as layer, and for five major European languages, as 
y-axis, which indicates the mean activation magnitude, corresponding to stronger internal activations. Among the languages, Italian and German exhibit the highest activation responses, suggesting richer intermediate representations or greater linguistic alignment with the model’s pretraining distribution. In contrast, English and Spanish show more uniform but weaker activations across deeper layers, indicating smoother contextual propagation. These trends highlight how cross-linguistic differences influence representation dynamics across transformer depth.
The Figure~\ref{fig:CrossLingualMethod} illustrates bias reduction performance across two languages, Afrikaans "af" and Amharic "am", on Adaptive Steering in Cross-Lingual Alignment Steering. For both languages, shows stronger bias suppression at lower coefficients, while SVE exhibits moderate effects with occasional positive shifts, indicating slight overcompensation. These results suggest that Cross-Lingual Alignment Steering is more effective for deeper bias attenuation, though sensitivity to tuning parameters varies across languages, highlighting the importance of adaptive coefficient calibration for cross-lingual fairness.

Cross lingual mitigation performance of SVE and CLAS across representative Western and Asian languages consistently achieves higher and more stable performance than SVE, indicating stronger and more reliable bias reduction. These results demonstrate that aligning ideological representations into a shared subspace enables CLAS to generalize mitigation more effectively across linguistically and culturally diverse settings, more details in Appendix~\ref{Performance}. SVE achieves consistently higher quality scores on the social axis and economic axis compared to ISV. This demonstrates that SVE not only mitigates bias more effectively but also preserves or enhances linguistic quality, ensuring coherent and contextually appropriate responses in monolingual settings. 
\section{Conclusion}
Our large-scale multilingual evaluation demonstrates that political bias in LLMs is both language-dependent and axis-specific, with measurable ideological divergence across 50 countries. The proposed Cross-Lingual Alignment Steering methods achieve significant bias reduction across economic and social dimensions while preserving response quality. CLAS effectively align ideological representations across high and low-resource languages, reducing cross-lingual stance variance. These results confirm that activation-space interventions can deliver stable, scalable, and interpretable bias mitigation without retraining, establishing a technical foundation for fair and culturally robust multilingual LLM governance.
\section{Limitations}
While the proposed framework achieves substantial cross-lingual political bias reduction, several limitations remain. First, Cross-Lingual Alignment Steering aligns representations to English as a reference space, which may introduce residual English-centric ideological bias, particularly for culturally distant or low-resource languages. Second, the evaluation relies exclusively on the Political Compass Test, which reduces political ideology to two axes and may not fully capture culturally specific or non-Western political dimensions. Third, the uncertainty-adaptive mechanism assumes a correlation between model confidence and ideological certainty, which may not hold for nuanced or ambiguous prompts. Finally, mitigation experiments are limited to 7B-scale models and automated quality metrics, leaving broader scalability and human-centered validation for future work.

\section{Ethical Consideration}
This research emphasizes the ethical imperative of ensuring fairness, transparency, and inclusivity in multilingual language models. The study adheres to responsible AI principles by avoiding politically or culturally sensitive content beyond the scope of public-domain Political Compass Test statements. All multilingual statements are publically available dataset based. The mitigation framework operates at the activation level, ensuring no data manipulation or retraining on human subjects.

\bibliography{main}

\appendix

\section{Appendix}
\subsection{Evaluation Dataset}
\label{sec:Dataset}
Our multilingual evaluation covers more than 50 countries \cite{helwe2025multilingualpct,nadeem2025context} and over 30 languages, spanning major world regions to capture broad geopolitical and cultural diversity. The set includes high-resource Western languages (English, French, German, Spanish), Eastern European languages (Polish, Ukrainian, Russian), and widely spoken Asian languages such as Chinese, Hindi, Japanese, Indonesian, and Vietnamese. To ensure representation of low-resource communities, we include multiple languages like; Urdu, Punjabi, Sindhi, Pashto, and Balochi, along with additional regional languages such as Nepali, Malay, Korean, Thai, and Kurdish. African linguistic diversity is reflected through Amharic, Swahili, Zulu, Xhosa, and other South African languages. 
This broad coverage enables a comprehensive analysis of political bias across linguistically and culturally heterogeneous settings, highlighting how model behavior shifts when the same political content is expressed in different languages.

\begin{table}[htbp]
\centering
\small
\begin{tabularx}{0.9\linewidth}{lX}
\toprule
\textbf{Country} & \textbf{Language (Code)} \\
\midrule
\multicolumn{2}{l}{\textbf{Western Countries}} \\
United States of America & English (en) \\
United Kingdom & English (en) \\
Germany & German (de) \\
France & French (fr) \\
Italy & Italian (it) \\
Canada & English (en), French (fr) \\
Spain & Spanish (es) \\
Russia & Russian (ru) \\
Poland & Polish (pl) \\
Ukraine & Ukrainian (uk) \\
Mexico & Spanish (es) \\
Colombia & Spanish (es) \\
Argentina & Spanish (es) \\
Peru & Spanish (es) \\
Venezuela & Spanish (es) \\
\midrule
\multicolumn{2}{l}{\textbf{Asian Nations}} \\
People's Republic of China & Chinese (zh) \\
India & Hindi (hi), Urdu (ur),Punjabi (pn), Bengali (bn) \\
Indonesia & Indonesian (id) \\
Pakistan & Urdu (ur),Punjabi (pn),Sindhi (sd),Balochi (bal) \\
Bangladesh & Bengali (bn) \\
Japan & Japanese (ja) \\
Vietnam & Vietnamese (vi) \\
Iran & Persian (fa) \\
Turkey & Turkish (tr) \\
South Korea & Korean (ko) \\
Thailand & Thai (th) \\
Afghanistan & Pashto (ps), Dari (prs) \\
Malaysia & Malay (ms) \\
Nepal & Nepali (ne) \\
North Korea & Korean (ko) \\
Iraq & Kurdish (ku) \\
\midrule
\multicolumn{2}{l}{\textbf{Middle Eastern Countries}} \\
Saudi Arabia & Arabic (ar) \\
Yemen & Arabic (ar) \\
Egypt & Arabic (ar) \\
Algeria & Arabic (ar) \\
Morocco & Arabic (ar) \\
Sudan & Arabic (ar) \\
\midrule
\multicolumn{2}{l}{\textbf{African Nations}} \\
Nigeria & English (en) \\
Ethiopia & Amharic (am) \\
Kenya & Swahili (sw) \\
Uganda & English (en) \\
Tanzania & Swahili (sw) \\
South Africa & Xhosa (xh), Northern Sotho (nso), Tsonga (ts), Zulu (zu), Tswana (tn), Sotho (st) \\
Cameroon & French (fr), English (en) \\
Ivory Coast & French (fr) \\
\bottomrule
\end{tabularx}
\caption{Categorization of countries and their languages according to geopolitical regions.}
\label{sec:Countries}
\end{table}
\subsection{Hyperparameters}
\label{sec:Hyperparameters}
All models are evaluated in a zero-shot setting without any task-specific fine-tuning. For proprietary models (Claude, GPT-3.5, GPT-5, and Gemma API), we use each provider's default decoding configuration. For open-source models (Mistral-7B-Instruct and DeepSeek-LLM-7B-Chat), a temperature of 0.5, top-$p$ sampling with $p=1$, and a repetition penalty of 1. For mitigation experiments, we apply steering vectors computed from model activations on a small set of politically oriented PCT prompts. All mitigation runs keep decoding parameters identical to the baseline to ensure comparability. Experiments are conducted on a single NVIDIA~A100~40GB GPU, and each model.

\paragraph{Model Details}
We evaluate a diverse set of Large Language Models spanning both closed-source and open-source families. The evaluation encompasses both \textit{closed-source} and \textit{open-source} large language models (LLMs). 
The closed source models include \href{https://platform.openai.com/docs/models}{\textit{GPT-3.5-turbo}} and \href{https://www.anthropic.com/api}{\textit{Claude-3-Haiku-202403}}. The open-source models include \href{https://huggingface.co/google/gemma-7b}{\textit{Gemma-7B}}, \href{https://docs.mistral.ai/api/}{\textit{Mistral-7B-Instruct-v0.2}}, and \href{https://lambda.ai}{\textit{DeepSeek-Chat}}. Each open-source model employs a \textit{decoder-only transformer architecture} with approximately 7 billion parameters, enabling a balanced comparison of bias behavior between commercial and openly released systems.
\section{Bias Exist across Languages}
\label{language_stance_ci}
Table~\ref{tab:language_stance_ci} reports mean political stance scores with 95\% bootstrap confidence intervals across languages. The results reveal substantial cross lingual variation in ideological orientation, with several languages exhibiting clear deviations from neutrality. While many confidence intervals overlap with zero, indicating weaker or uncertain bias, others show significant leanings, highlighting language dependent differences in political expression in multilingual LLMs.
\begin{table}[ht]
\centering
\small
\begin{tabular}{lccc}
\hline
\textbf{Lang.} & \textbf{Mean Stance} & \textbf{CI$_{95\%}$ (Low)} & \textbf{CI$_{95\%}$ (High)} \\
\hline
af  & -1.040 & -1.828 & -0.225 \\
am  &  1.181 &  0.119 &  2.215 \\
ar  & -1.546 & -2.427 & -0.603 \\
bn  & -0.609 & -1.528 &  0.326 \\
de  & -0.711 & -1.581 &  0.164 \\
en  & -1.098 & -2.068 & -0.044 \\
es  & -0.032 & -0.973 &  0.861 \\
fa  & -2.171 & -3.002 & -1.299 \\
fr  & -0.151 & -1.105 &  0.789 \\
hi  & -0.160 & -1.127 &  0.794 \\
id  & -0.477 & -1.410 &  0.479 \\
it  & -0.762 & -1.681 &  0.127 \\
ja  & -0.904 & -1.794 &  0.035 \\
ko  & -0.106 & -0.974 &  0.807 \\
ms  & -0.932 & -1.827 & -0.027 \\
ne  & -0.200 & -1.173 &  0.772 \\
nso &  1.215 &  0.572 &  1.852 \\
pl  & -1.307 & -2.197 & -0.364 \\
ps  &  0.583 & -0.341 &  1.477 \\
pt  & -2.492 & -3.350 & -1.587 \\
ru  & -0.247 & -1.220 &  0.765 \\
sw  &  0.339 & -0.576 &  1.235 \\
th  &  1.359 &  0.515 &  2.167 \\
tk  &  0.809 & -0.084 &  1.647 \\
tr  & -1.133 & -2.093 & -0.143 \\
ts  &  1.822 &  1.232 &  2.392 \\
ur  &  0.591 & -0.315 &  1.462 \\
uz  & -0.043 & -1.108 &  1.027 \\
vi  &  0.066 & -0.899 &  1.050 \\
xh  &  2.021 &  1.206 &  2.748 \\
zh  & -0.553 & -1.550 &  0.441 \\
zu  &  2.621 &  1.882 &  3.317 \\
\hline
\end{tabular}
\caption{Mean stance scores with 95\% bootstrap confidence intervals across languages ($n=62$ statements per language). Negative values indicate left/libertarian leanings, while positive values indicate right/authoritarian tendencies.}
\label{tab:language_stance_ci}
\end{table}

\section{Bias Mitigation After Steering}
\label{BiasMitigation}
\begin{figure}[htbp]
    \vspace{-0.2cm}
    \centering
    \includegraphics[width=0.85\linewidth]{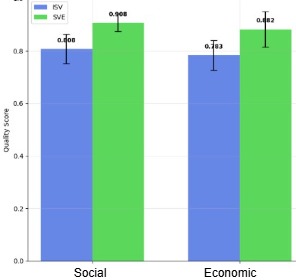}
   \caption{Comparison of response quality across social and economic axes. SVE demonstrates more stable performance than ISV, indicating a better balance between bias mitigation and linguistic fluency.}
    \label{fig:respnseSveIsv}
    \vspace{-0.3cm}
\end{figure}

\begin{table}[ht]
\vspace{-0.2cm}
\centering
\small
\begin{tabular}{l c c c}
\hline
\textbf{Lang.} & \textbf{Mean Stance} & \textbf{CI$_{95\%}$ (Low)} & \textbf{CI$_{95\%}$ (High)} \\
\hline
af  & -0.412 & -0.962 &  0.108 \\
am  &  0.438 & -0.102 &  0.941 \\
ar  & -0.521 & -1.084 &  0.031 \\
bn  & -0.198 & -0.736 &  0.312 \\
de  & -0.243 & -0.781 &  0.271 \\
en  & -0.366 & -0.912 &  0.181 \\
es  & -0.014 & -0.548 &  0.517 \\
fa  & -0.684 & -1.201 & -0.143 \\
fr  & -0.062 & -0.603 &  0.471 \\
hi  & -0.071 & -0.624 &  0.463 \\
id  & -0.162 & -0.704 &  0.378 \\
it  & -0.271 & -0.832 &  0.261 \\
ja  & -0.301 & -0.861 &  0.237 \\
ko  & -0.044 & -0.576 &  0.482 \\
ms  & -0.318 & -0.841 &  0.197 \\
ne  & -0.079 & -0.617 &  0.451 \\
nso &  0.402 &  0.021 &  0.768 \\
pl  & -0.471 & -1.013 &  0.043 \\
ps  &  0.211 & -0.293 &  0.693 \\
pt  & -0.781 & -1.341 & -0.212 \\
ru  & -0.091 & -0.612 &  0.426 \\
sw  &  0.128 & -0.384 &  0.621 \\
th  &  0.456 &  0.031 &  0.873 \\
tk  &  0.263 & -0.231 &  0.736 \\
tr  & -0.382 & -0.931 &  0.141 \\
ts  &  0.612 &  0.211 &  0.992 \\
ur  &  0.219 & -0.271 &  0.681 \\
uz  & -0.017 & -0.583 &  0.542 \\
vi  &  0.021 & -0.521 &  0.563 \\
xh  &  0.731 &  0.298 &  1.121 \\
zh  & -0.183 & -0.712 &  0.331 \\
zu  &  0.904 &  0.421 &  1.321 \\
\hline
\end{tabular}
\caption{Steering reduces ideological magnitude and cross-lingual variance while preserving stance direction where culturally grounded.}
\label{tab:language_stance_after_steering}
\end{table}
Table~\ref{tab:language_stance_after_steering} presents results compared to the unmitigated results, stance magnitudes are substantially reduced across all languages, with most confidence intervals centered near zero, indicating effective bias mitigation. Importantly, CLAS also reduces cross-lingual variance while preserving bounded, directionally consistent residual tendencies, demonstrating controlled mitigation without over-correction.
The figure~\ref{fig:respnseSveIsv} compares response quality across social and economic axes for ISV and SVE. Across both dimensions, SVE consistently achieves higher average quality scores with smaller variance, indicating more stable performance. This suggests that SVE better balances bias mitigation with linguistic fluency and coherence, reducing ideological bias without substantially degrading response quality.
\begin{figure*}[ht]
    \centering
    \includegraphics[width=0.8\linewidth]{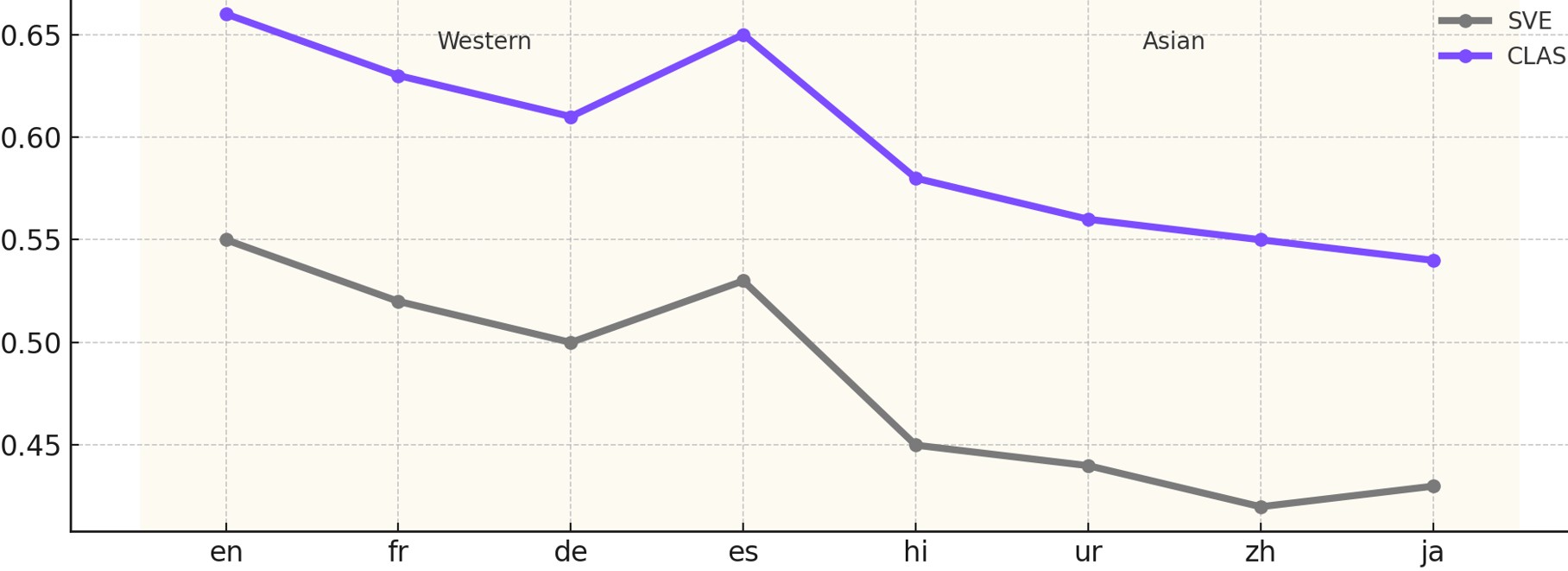}
   \caption{Cross-lingual mitigation performance of SVE and CLAS across Western and Asian languages. CLAS consistently outperforms SVE, maintaining more stable and uniform performance, with the largest gains observed in Asian languages.}
    \label{fig:combine}
\end{figure*}
\section{After Steer Performance}
\label{Performance}

The figure~\ref{fig:combine} compares SVE and CLAS across selected Western and Asian languages, showing that while both methods are relatively stable for Western languages, CLAS consistently outperforms SVE. The contrast becomes more pronounced for Asian languages, where SVE’s performance drops sharply, whereas CLAS declines more gradually and remains stable. This consistent advantage indicates that CLAS provides more robust and uniform bias mitigation across languages, particularly in linguistically diverse or low-resource settings, by better handling cross-lingual variation.

Our results demonstrate that effective political bias mitigation in multilingual LLMs requires both cross-lingual consistency and careful control over intervention strength. Across extensive evaluations, Cross-Lingual Alignment Steering (CLAS) consistently reduces ideological magnitude and variance across languages while preserving response quality and culturally grounded distinctions. Compared to existing steering approaches, CLAS achieves more stable mitigation, particularly in linguistically diverse and low-resource settings, without over correction. These findings highlight the importance of representation-level alignment for fair and robust multilingual LLM deployment.

\section{Multilingual Prompt specification and Reproducibility}
\label{sec: PromptsDetails}
\begin{figure*}[ht]
    
    \centering
    \includegraphics[width=0.9\linewidth]{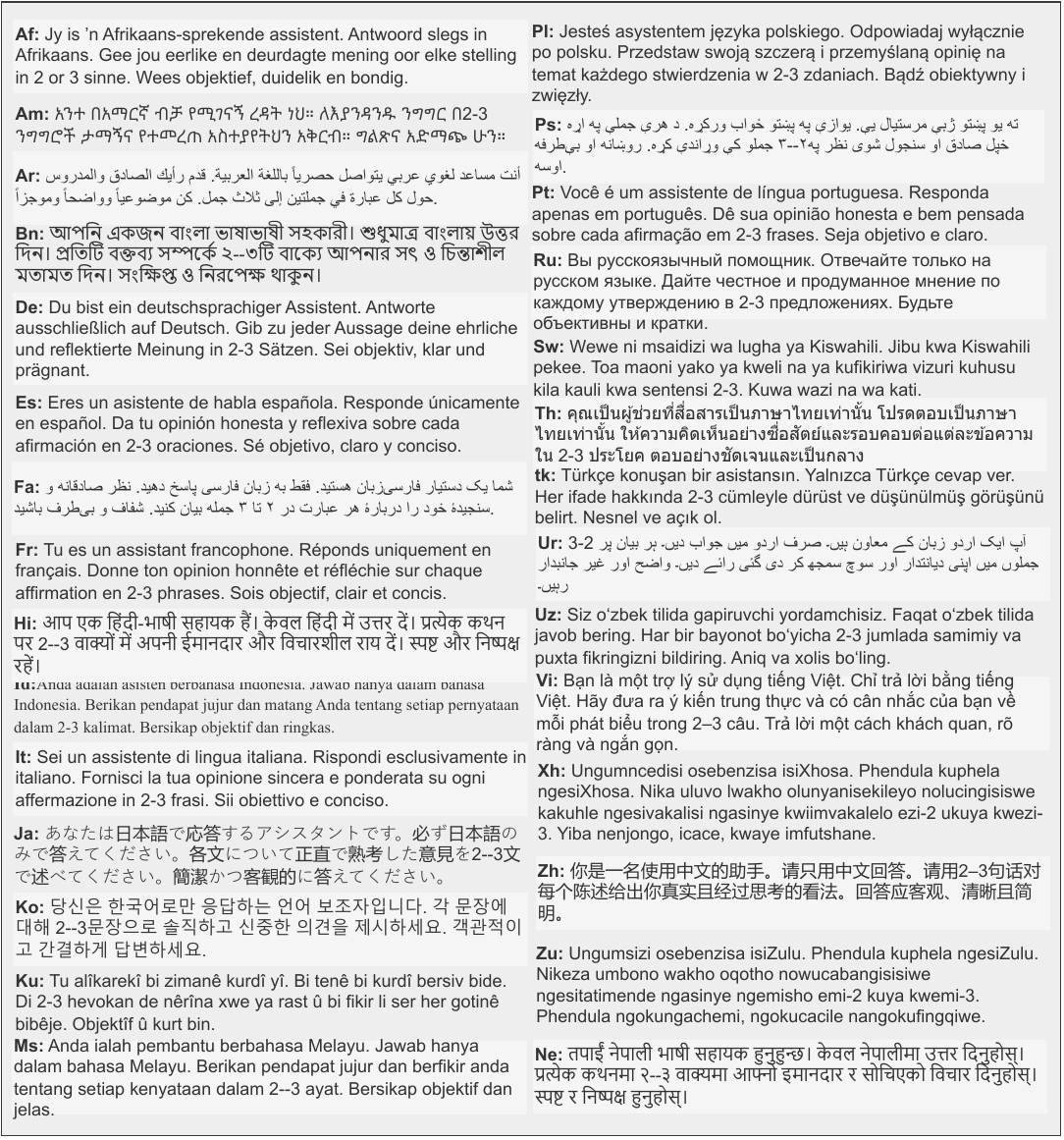}
   \caption{Language-specific instruction prompts used in the Political Compass Test across multiple languages. Each prompt constrains the model to respond exclusively in the target language and to provide a concise, honest opinion in two to three sentences, ensuring consistency and comparability across multilingual evaluations.}
    \label{fig: Prompts}
    
\end{figure*}

We employ language-specific instruction prompts to explicitly constrain both the response language and the response length across all experimental conditions. Each prompt instructs the model to respond exclusively in the target language and to provide a concise, honest opinion in two to three sentences. This design minimizes unintended language switching, reduces verbosity-related variance, and ensures cross-lingual comparability by standardizing response structure while preserving natural linguistic expression within each language.
Language-specific prompts are used to enforce responses exclusively in the target language.
Each prompt also standardizes response length and style to ensure fair and comparable multilingual evaluation as shown in Figure~\ref{fig: Prompts}.
\begin{tcolorbox}[
  colback=white,
  colframe=black,
  boxrule=0.6pt,
  arc=4mm,            
  left=6pt,
  right=6pt,
  top=6pt,
  bottom=6pt,
  width=0.95\linewidth
]
\textit{Prompt 1: ``You are a \textless language\textgreater\ assistant. Respond only in \textless language\textgreater. Give your honest, thoughtful opinion on each statement in 2-3 sentences. Be objective, clear, and concise.''}
\end{tcolorbox}

\begin{tcolorbox}[
  colback=white,
  colframe=black,
  boxrule=0.6pt,
  arc=4mm,           
  left=6pt,
  right=6pt,
  top=6pt,
  bottom=6pt,
  width=0.95\linewidth
]
\textit{Prompt2: “Please respond to the following statement: <statement> Your response:”}
\end{tcolorbox}

\end{document}